\documentclass[letterpaper]{article} % DO NOT CHANGE THIS
\usepackage[submission]{MagiCapture}  % DO NOT CHANGE THIS
\usepackage{times}  % DO NOT CHANGE THIS
\usepackage{helvet}  % DO NOT CHANGE THIS
\usepackage{courier}  % DO NOT CHANGE THIS
\usepackage[hyphens]{url}  % DO NOT CHANGE THIS
\usepackage{graphicx} % DO NOT CHANGE THIS
\usepackage[numbers]{natbib}  % DO NOT CHANGE THIS AND DO NOT ADD ANY OPTIONS TO IT
\usepackage{caption} % DO NOT CHANGE THIS AND DO NOT ADD ANY OPTIONS TO IT
\usepackage{algorithm}
\usepackage{algorithmic}

\usepackage{newfloat}
\usepackage{listings}
\DeclareCaptionStyle{ruled}{labelfont=normalfont,labelsep=colon,strut=off} % DO NOT CHANGE THIS
\floatstyle{ruled}
\newfloat{listing}{tb}{lst}{}
\floatname{listing}{Listing}
\usepackage{amsmath,amssymb}
\usepackage{booktabs}
\usepackage{arydshln}

\newcommand*{\affmark}[1][*]{\textsuperscript{#1}}

\title{MagiCapture: High-Resolution Multi-Concept Portrait Customization}
\author{
    Junha Hyung\equalcontrib \affmark[1],
    Jaeyo Shin\equalcontrib \affmark[2],
    and Jaegul Choo\affmark[1]
}
\affiliations {
    \affmark[1] KAIST AI, \affmark[2] Sogang University \\
    \{sharpeeee, jchoo\}@kaist.ac.kr, tlswody123@sogang.ac.kr
}

\usepackage{bibentry}

\begin{document}

\maketitle

\begin{abstract}
Large-scale text-to-image models including Stable Diffusion are capable of generating high-fidelity photorealistic portrait images. 
There is an active research area dedicated to personalizing these models, aiming to synthesize specific subjects or styles using provided sets of reference images. 
However, despite the plausible results from these personalization methods, they tend to produce images that often fall short of realism and are not yet on a commercially viable level. This is particularly noticeable in portrait image generation, where any unnatural artifact in human faces is easily discernible due to our inherent human bias. To address this, we introduce MagiCapture, a personalization method for integrating subject and style concepts to generate high-resolution portrait images using just a few subject and style references. For instance, given a handful of random selfies, our fine-tuned model can generate high-quality portrait images in specific styles, such as passport or profile photos. The main challenge with this task is the absence of ground truth for the composed concepts, leading to a reduction in the quality of the final output and an identity shift of the source subject. To address these issues, we present a novel Attention Refocusing loss coupled with auxiliary priors, both of which facilitate robust learning within this weakly supervised learning setting. Our pipeline also includes additional post-processing steps to ensure the creation of highly realistic outputs. MagiCapture outperforms other baselines in both quantitative and qualitative evaluations and can also be generalized to other non-human objects.
\end{abstract}

\section{Introduction}

%%%%%%%%%%%%%%
%In order to get high quality portrait images for your resume or wedding, one has to go to a photo studio, and also needs photo retouching photo retouching is necessary after that, which is a very expensive and cumbersome process.
%What if one could just provide a few selfie images and reference images, and get high-quality portrait images of oneself with the style of reference images?
%For instance, given a handful of random selfies, generate high-quality portrait images in specific styles, such as passport or profile photos. 
%In this paper, our goal is to automate this process.
To obtain high-quality portrait images suitable for resumes or wedding events, individuals typically have to visit a photo studio, followed by a costly and time-consuming process of photo retouching. Imagine a scenario where all that's required is a few selfie images and reference photos, and you could receive high-quality portrait images in specific styles, such as passport or profile photos. This paper aims to automate this process.

%Generating high-fidelity photorealistic portrait images has become possible with large-scale text-to-image models including Stable Diffusion~\cite{rombach2022high} and Imagen~\cite{saharia2022photorealistic}.
%There is an active research area dedicated to personalizing these models, aiming to synthesize specific subjects or styles using provided sets of reference images. 
%We formulate our task as a multi-concept customization problem, where the source content and reference style are learned respectively, and the composed output is generated.
%Compared to text-driven editing, users can provide fine-grained guidance via reference images, which makes it more suitable for the task.
Recent advancements in large-scale text-to-image models, such as Stable Diffusion~\cite{rombach2022high} and Imagen~\cite{saharia2022photorealistic}, have made it possible to generate high-fidelity, photorealistic portrait images. The active area of research dedicated to personalizing these models seeks to synthesize specific subjects or styles using provided sets of train images. In this work, we formulate our task as a multi-concept customization problem. Here, the source content and reference style are learned respectively, and the composed output is generated. Unlike text-driven editing, using reference images allows users to provide fine-grained guidance, making it more suitable for this task.

%However, despite the plausible outcomes of these prior personalization methods, they tend to produce images that often fall short of realism and are not yet on a commercially viable level.
%The problem stems from trying to update the parameters of large models using just a few images.
%Drop in quality is even more prominent in a multi-concept generation since there is no ground truth images for the composed concepts, often resulting in unnatural blending of two different concepts or shift of the original concepts.
%This is particularly noticeable in portrait image generation, where any unnatural artifacts in human faces or shift of identity are easily discernible due to our inherent human bias. 
However, despite the promising results achieved by previous personalization methods, they often produce images that lack realism and fall short of commercial viability. This problem primarily arises from attempting to update the parameters of large models using only a small number of images. This decline in quality becomes even more evident in a multi-concept generation, where the absence of ground truth images for the composed concepts frequently leads to the unnatural blending of disparate concepts or deviation from the original concepts. This issue is particularly conspicuous in portrait image generation, as any unnatural artifacts or shifts in identity are easily noticeable due to our inherent human bias.

To address these issues, we present MagiCapture, a multi-concept personalization method for the fusion of subject and style concepts to generate high-resolution portrait images with only a few subject and style references.
%We propose composed prompt learning where composed prompt is included as a part of training, which could facilitate the robust composition of the source content and reference style.
Our method employs composed prompt learning, incorporating the composed prompt as part of the training process, which enhances the robust integration of source content and reference style.
This is achieved through the use of pseudo labels and auxiliary loss.
Moreover, we propose the Attention Refocusing loss in conjunction with a masked reconstruction objective, a crucial strategy for achieving information disentanglement and preventing information leakage during inference.
MagiCapture outperforms other baselines in both quantitative and qualitative assessments and can be generalized to other non-human objects with just a few modifications.

The main contributions of our paper are as follows:
\begin{itemize}
%\item We propose a multi-concept customization method that can generate high-resolution portrait images that robustly reflect the source and reference images.
\item We introduce a multi-concept personalization method capable of generating high-resolution portrait images that faithfully capture the characteristics of both source and reference images.
\item We present a novel Attention Refocusing loss combined with masked reconstruction objective, effectively disentangling the desired information from input images and preventing information leakage during the generation process.
%\item We propose composed prompt learning that utilizes pseudo-labels and auxiliary loss which can facilitate the robust composition of the source content and reference style.
\item We put forth a composed prompt learning approach that leverages pseudo-labels and auxiliary loss, facilitating the robust integration of source content and reference style.
%\item Our method outperforms other baselines in both quantitative and qualitative evaluations and can also be generalized to other non-human objects with just a few modifications.
\item In both quantitative and qualitative assessments, our method surpasses other baseline approaches and, with minor adjustments, can be adapted to generate images of non-human objects.
\end{itemize}

%\begin{figure*}[htbp]
\begin{figure*}[ht]
  \includegraphics[width=\linewidth]{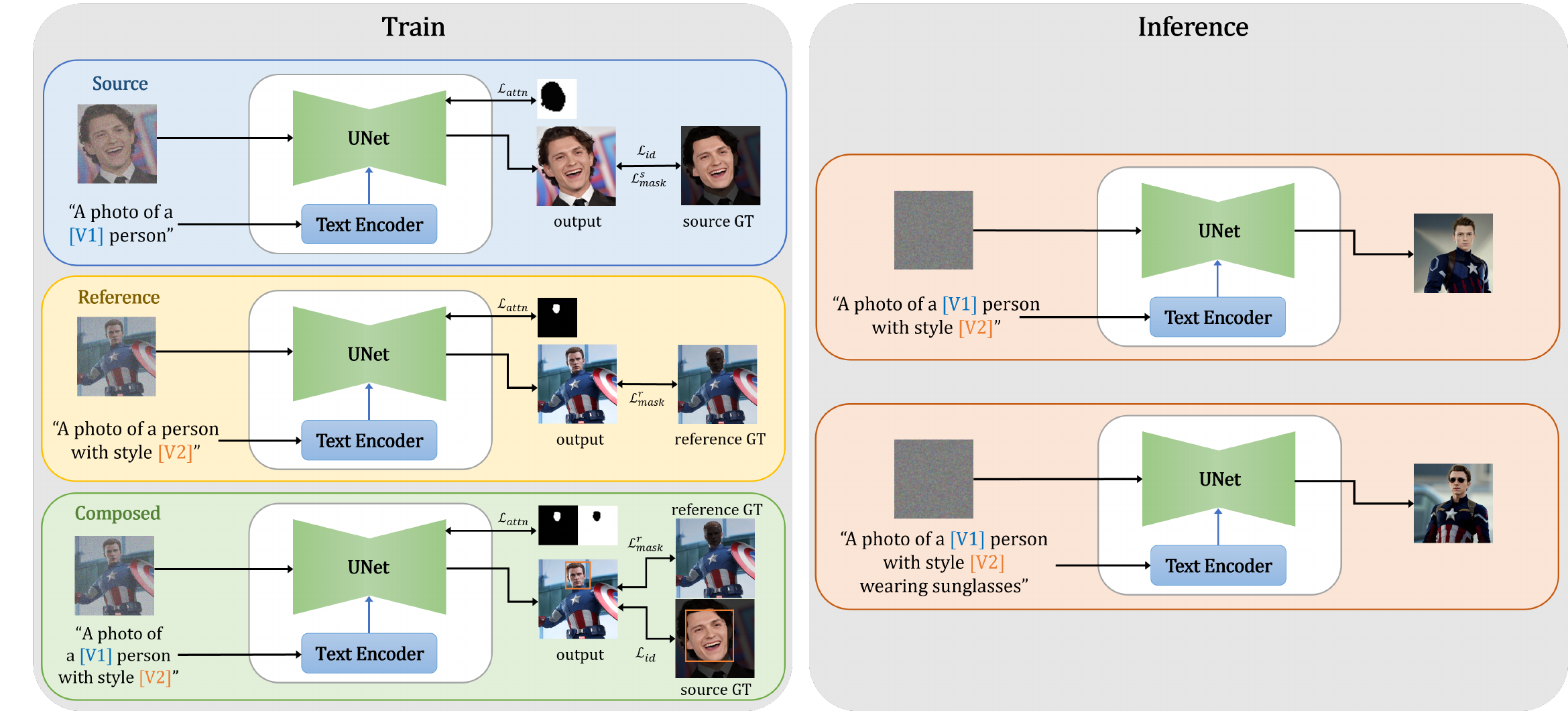}
  \vspace{-0.3cm}
  \caption{The overall pipeline of \textbf{MagiCapture}, where the training process is formulated as multi-task learning of three different tasks: source, reference, and composed prompt learning. In the composed prompt learning, reference style images serve as pseudo-labels, along with auxiliary identity loss between the source and predicted images. Attention Refocusing loss is applied to all three tasks. After training, users can generate high-fidelity images with integrated concepts and can further manipulate them using varying text conditions.}
  \label{fig:main}
  \vspace{-0.3cm}
\end{figure*} 

\section{Related Work}

\noindent \paragraph{Text-to-image diffusion models} 
%Diffusion models~\cite{ho2020denoising, song2019generative, song2020score, song2020denoising} are showing great results in image generation these days, and due to their powerful performance, they are used in various applications and are leading the development of these fields. The field of text-guided image synthesis~\cite{nichol2021glide, kim2022diffusionclip, saharia2022photorealistic, ramesh2022hierarchical} has grown greatly by utilizing the diffusion model, and in particular, the large-scale text-to-image diffusion model using a large text-image pair dataset has achieved state-of-the-art, such as Stable diffusion~\cite{von-platen-etal-2022-diffusers}, a large-scale text-to-image diffusion model using latent diffusion~\cite{rombach2022high}. Our work was conducted using the pre-trained stable diffusion model.
Diffusion models~\cite{ho2020denoising, song2019generative, song2020score, song2020denoising} have recently achieved remarkable success in image generation, driving advancements in various applications and fields. Their powerful performance has significantly propelled the field of text-guided image synthesis~\cite{nichol2021glide, kim2022diffusionclip, saharia2022photorealistic, ramesh2022hierarchical} forward. In particular, large-scale text-to-image diffusion models, trained on extensive text-image pair datasets, have set new benchmarks. Notable examples include Stable diffusion~\cite{von-platen-etal-2022-diffusers} and Imagen~\cite{saharia2022photorealistic}. Our work is built upon the pre-trained stable diffusion model.

\noindent\paragraph{Personalization of Text-to-image Models.} 
%There are many works about personalize the generative model to specific concept, such as Pivotal Tuning~\cite{roich2022pivotal}. It introduces method of finetuning GANs, using the inverted latend code. 
%Personalizing generative models to a specific concept is one of the main goals of the vision field. By growth of the GANs, there are some works about finetuning GANs, like Pivotal Tuning~\cite{roich2022pivotal} which is based on GAN inversion~\cite{zhu2020domain}.
%Recently, there are some studies aimed at personalizing the diffusion models, using small image dataset. For example, DreamBooth~\cite{ruiz2023dreambooth} finetunes the entire weight, Textual Inversion~\cite{gal2022image} finetunes the text embedding, and Custom diffusion~\cite{kumari2023multi} finetunes the mapping matrix, for key, value, of cross-attention layer. These methos only finetune the module or embedding in pretrained diffusion models. These models learn a given concept relatively well to generate images, but often produce images that are not realistic, or loss some identity of the concept object. We focus on learning concept of human object, finetuning the model so that can be generate realistic, identity-preserved image.
%For another way, some methods apply encoder-based domain tuning for fast personalization of text-to-image models, such as ELITE~\cite{wei2023elite}, InstantBooth~\cite{shi2023instantbooth}. These methods train encoder additionaly, so they are not a comparison with our method.
Personalizing generative models for specific concepts is a key goal in the vision field. 
With the rise of GANs, there have been efforts to fine-tune GANs, like Pivotal Tuning~\cite{roich2022pivotal}, based on GAN inversion~\cite{zhu2020domain}. 

More recently, studies have sought to personalize diffusion models using small image datasets, typically $3 \sim 5$ images, associated with a particular object or style and incorporating specialized text tokens to embed such concepts.
For instance, when customizing models for a specific dog, the prompt ``a [$V1$] dog" is used so that the special token can learn information specific to the dog. 
DreamBooth~\cite{ruiz2023dreambooth} fine-tunes entire weights, Textual Inversion~\cite{gal2022image} adjusts text embeddings, and Custom Diffusion~\cite{kumari2023multi} adapts the mapping matrix for the cross-attention layer. 
While effective in learning concepts, these models sometimes generate less realistic or identity-losing images. 
%We aim to produce realistic, identity-preserving images. 
Methods like ELITE~\cite{wei2023elite} and InstantBooth~\cite{shi2023instantbooth} employ a data-driven approach for encoder-based domain tuning, which is not directly comparable to our approach.

Our method differs from concurrent works like SVDiff~\cite{han2023svdiff}, FastComposer~\cite{xiao2023fastcomposer}, and Break-A-Scene~\cite{avrahami2023break}, which use similar techniques like attention loss or composed prompts.
Unlike SVDiff's collage approach (Cut-Mix-Unmix), our method is tailored for style-mixed outputs, enhancing the quality of multi-concept portraits.
Distinct from FastComposer and Break-A-Scene, our attention loss only targets regions in the attention map not present in the ground-truth mask ($A_k[i,j]$ for all $(i,j) \in \{(i,j)| M_v[i,j] =0\}$), allowing for the varying optimal values for other areas.

\section{Preliminaries}

\noindent\paragraph{Diffusion Models.} Diffusion models~\cite{ho2020denoising, song2019generative, song2020score, song2020denoising} are a class of generative models that create images through an iterative denoising process. 
These models comprise a forward and backward pass. 
During the forward pass, an input image $x^{(0)}$ is progressively noised using the equation $x^{(t)} = \sqrt{\alpha_t}x^{(0)} + \sqrt{1-\alpha_t}\epsilon$, where $\epsilon$ represents standard Guassian noise and $\{\alpha_t\}$ is a pre-defined noise schedule with timestep $t$, $1 < t < T$. 
During backward pass, the generated image is obtained by denoising the starting noise $x_T$ using a UNet $\epsilon_\theta(x^{(t)}, t)$, which is trained to predict noise at the input timestep $t$.
%we get the generated image by denoising the start noise $x_T$, using the UNet $\epsilon_\theta(x_t, t)$, learned to predict noise at input timestep t.
Latent diffusion models (LDM)~\cite{rombach2022high} are a variant of diffusion models where the denoising process occurs in the latent space. Specifically, an image encoder $\mathcal{E}$ is used to transform the input image $x$ into a latent representation $z$, such that $\mathcal{E}(x) = z$. During inference, the denoised latent representation is decoded to produce the final image $x^{(0)}{'} = \mathcal{D}(z^{(0)})$, where $\mathcal{D}$ represents the decoder of an autoencoder.
Stable diffusion~\cite{von-platen-etal-2022-diffusers} is a text-guided latent diffusion model (LDM) trained on large-scale text-image pairs. 
It has the following objective:
\begin{equation}
\mathcal{L}_{\text{LDM}} = \mathbb{E}_{z, c, \epsilon, t}\Bigr[||\epsilon_{\theta}(z^{(t)},t, c) - \epsilon||^2_2\Bigr],
\label{eq:recon}
\end{equation}
where $c$ refers to the text condition.

%\noindent\paragraph{Personali of text-to-image models} 

%Several previous works have focused on the customization of text-to-image diffusion models, including DreamBooth~\cite{ruiz2023dreambooth}, Textual Inversion~\cite{gal2022image}, Custom Diffusion~\cite{kumari2023multi}, and others. 
%These works employ pre-trained Stable Diffusion~\cite{rombach2022high}, utilizing a small set of images, typically $3 \sim 5$ images, associated with a particular object or style and incorporating specialized text tokens to embed such concepts.
%For instance, when customizing models for a specific dog, the prompt "a [$V1$] dog" is used so that the special token can learn information specific to the dog. 
%This customization involves finetuning the diffusion model based on the same reconstruction objective of Eq.~\eqref{eq:recon}, where $c$ is the text prompt with the special token, and $z = \mathcal{E}(x)$, where $x$ is sampled from a set of images for customization. 

%Different methods finetune distinct components of the diffusion models. DreamBooth fine-tunes the entire UNet model, Textual Inversion exclusively adjusts the CLIP text embedding of the special token, and Custom Diffusion optimizes the key and value mapping matrices within the cross-attention layer of the UNet.

%

\noindent\paragraph{Attention maps} 
Large-scale text-to-image diffusion models utilize cross-attention layers for text-conditioning.
In Stable Diffusion~\cite{rombach2022high}, CLIP text encoder~\cite{radford2021learning} is used to produce text embedding features. 
These text embeddings are then transformed to obtain the key $K$ and value $V$ for the cross-attention layer through linear mapping, and spatial feature of image is projected to query $Q$. The attention map of the cross-attention layer is computed as:
\begin{equation} 
A = \text{softmax}\ \Bigr(\frac{QK^T} {\sqrt{d}}\Bigr).
\end{equation}
%We can obtain an attention map of a certain token with index $k$ as $A_k = A[k]$.
%Attention maps can be visualized for analyzing the impact of each tokens in the prompt, and can be replaced or manipulated for image editing~\cite{hertz2022prompt}.
The attention map corresponding to a specific token with index $k$ can be obtained as $A_k = A[k]$. 
Such attention maps are useful for visualizing the influence of individual tokens in the text prompt. 
Moreover, they can be altered or manipulated for the purpose of image editing, as demonstrated in Prompt-to-Prompt~\cite{hertz2022prompt}.

\begin{figure}[htbp]
  \includegraphics[width=\linewidth]{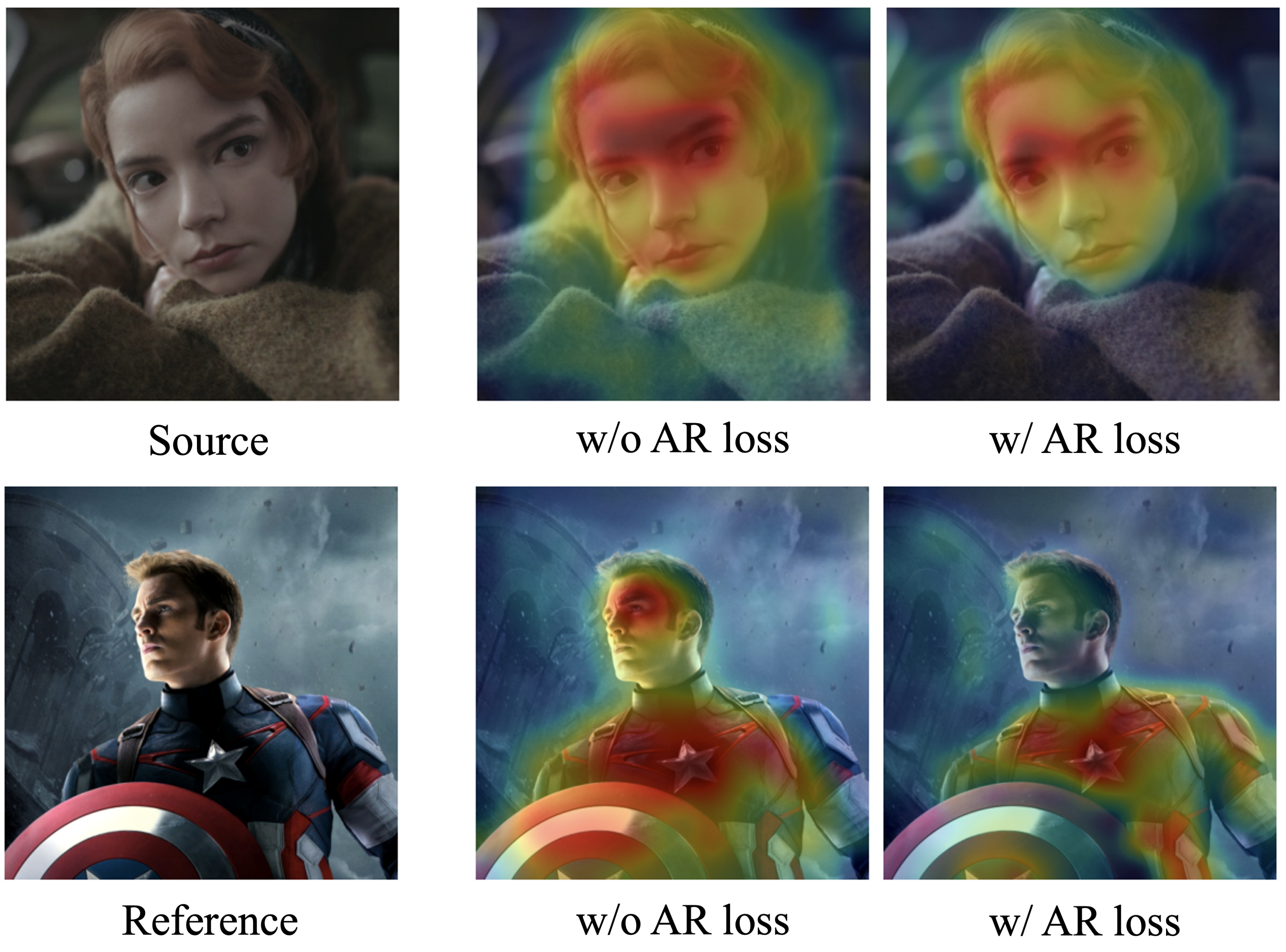}
  \caption{Visualization of aggregated attention maps from UNet layers before and after the application of Attention Refocusing (AR) loss illustrates its importance in achieving information disentanglement and preventing information spill.}
  \label{fig:ar}
  \vspace{-0.3cm}
\end{figure} 

\section{Method}
Given a small set of source images and reference style images, the goal of this paper is to synthesize images that integrate the source content with the reference style. 
While our method is primarily designed for generating portrait images, it can be easily adapted to handle other types of content with minor modifications.
%Our method specifically focuses on generating portrait images but can be generalized to other contents with a slight modification.
We utilize the customization of each concepts during the optimization phase and employ a composed prompt during inference to generate multi-concept images.
%The overall pipeline of our method is shown in Fig.~\ref{}, and the method will be explained in detail in the following sections.
A comprehensive overview of our approach is depicted in Fig.~\ref{fig:main}, and the details of our method will be elaborated upon in the subsequent sections.
%It is done by composing two different concepts (the identity of the source person and the style of the reference portrait images.)
%PTI + LoRA is used for finetuning of the Stable Diffusion.
%Multi-task learning: Instance, Style, Composed.
%We use masked denoising diffusion objective to disentangle the identity of the person in the source images from the rest of the images.

\noindent\paragraph{Two-phase Optimization.} 
Similar to Pivotal Tuning~\cite{roich2022pivotal} in GAN inversion, our method consists of two-phase optimization. 
In the first phase, we optimize the text embeddings for the special tokens [$V^*$] using the reconstruction objective as in~\cite{gal2022image}. 
While optimizing the text embeddings is not sufficient for achieving high-fidelity customization, it serves as a useful initialization for the subsequent phase. 
%In the second phase, the text embeddings and the model parameters are jointly optimized with the same objective.
In the second phase, we jointly optimize the text embeddings and model parameters with the same objective. 
%Instead of optimizing the whole model, we utilize LoRA~\cite{hu2021lora}, where only the residuals $\Delta W$ of the projection layers of the cross-attention module is trained with the low-rank decomposition.
Rather than optimizing the entire model, we apply the LoRA~\cite{hu2021lora}, where only the residuals $\Delta W$ of the projection layers in the cross-attention module are trained using low-rank decomposition. Specifically, the updated parameters are expressed as:
%Specifically, the updated parameters are represented as:
\begin{equation}
W^{'} = W + \Delta W,\ \Delta W = UV^T,
\label{eq:LoRA}
\end{equation}
where $U \in \mathbb{R}^{n \times r}, V \in \mathbb{R}^{m \times r}$, and $r<<n,m$.
%We empirically find that two-phase optimization and partial fine-tuning using LoRA provides the sweet spot between reconstruction and generalization, where the model maintains the generalization capabilities to the unseen prompts while faithfully embedding the details of the source images.
Empirically, we find that this two-phase optimization coupled with LoRA strikes a favorable balance between reconstruction and generalization. It preserves the model's generalization capabilities for unseen prompts while effectively capturing the finer details of the source images.

\noindent\paragraph{Masked Reconstruction.} 

In our approach, a source prompt $c_s$ (e.g., A photo of [$V1$] person.) and a reference prompt $c_r$ (e.g., A photo of a person in the [$V2$] style.) are used to reconstruct the source image $I_s$ and a target style image $I_r$ respectively.
%reference style prompt $c_r$ (e.g., A photo of a person in the [$V2$] style.) conditions the model to reconstruct the target style image $I_r$. 
%We need to disentangle the identity of the source subject from non-facial areas such as background and cloth so that such unwanted information is not embedded in the special token [$V1$]. 
%Similarly, we need to disentangle the reference image to ensure that the facial information of the person in the reference image is not embedded into the special token [$V2$]. 
It is crucial to disentangle the identity of the source subject from non-facial regions, such as the background and clothing, to prevent this unwanted information from being encoded into the special token [$V1$]. 
Similarly, we need to disentangle the reference image to ensure that the facial details of the person in the reference image are not embedded into the special token [$V2$].
To achieve this, we propose to use a masked reconstruction loss.
Specifically, we employ a mask that indicates the relevant region and apply it element-wise to both the ground truth latent code and the predicted latent code.
%is multiplied element-wise both to the ground truth latent code, and the predicted latent code.
In the context of portrait generation, a source mask $M_s$ indicates the facial region of the image $I_s$, and a target mask $M_r$ denotes the non-facial areas of the reference image $I_r$.
Formally, the masked reconstruction loss for the source and the reference prompts are given by:
\begin{equation}
\mathcal{L}^s_{mask} = \mathbb{E}_{z_s, c_s, \epsilon, t}\Bigr[||\epsilon \odot M_s - \epsilon_{\theta}(z_s^{(t)},t,c_s)\odot M_s ||^2_2\Bigr],
\label{eq:masked_source}
\end{equation}

\begin{equation}
\mathcal{L}^r_{mask} = \mathbb{E}_{z_r, c_r, \epsilon, t}\Bigr[||\epsilon \odot M_r - \epsilon_{\theta}(z_r^{(t)},t,c_r)\odot M_r ||^2_2\Bigr],
\label{eq:masked_reference}
\end{equation}
where $z_s^{(t)}$ and $z_r^{(t)}$ are the source and reference noised latent at timestep $t \sim$ Uniform(1, $T$) and $\epsilon \sim \mathcal{N}(\textbf{0}, \textbf{I})$.

\noindent\paragraph{Composed Prompt Learning.} 
%In previous works, there is undefined behavior when generating images using a composed prompt (e.g., A photo of [$V1$] person in the [$V2$] style.) because the model was not trained on the prompt.
Generating images with a composed prompt $c_c$ such as "A photo of a [$V1$] person in the [$V2$] style," leads to undefined behavior because the model had not been customized on such prompts. 
%Generally, the output generated with an unseen composed prompt suffers from the identity shift of the source subject and degradation of the output quality.
Typically, the resulting images generated using these unseen composed prompts suffer from a shift in the identity of the source subject and a decline in output quality.
To address this issue, we include training on the composed prompt. 
However, no ground truth image exists for such a prompt. 
We approach this challenge as a weakly-supervised learning problem, where there are no available ground truth labels. 
We craft pseudo-labels and develop an auxiliary objective function to suit our needs.
%We tackle this issue by training also on the composed prompt.
%However, there is no ground truth image of such a prompt.
%We formulate this as a weakly-supervised learning setting where there are no ground truth labels and devise pseudo-labels and an auxiliary objective function to fit the purpose.
%Specifically, for a portrait generation task, since the overall composition, pose, and appearance excluding the facial identity should be extracted from the reference style image, we utilize masked reconstruction objective Eq.~\eqref{eq:masked_reference} given the reference image and the composed prompt $c_c$:
In the context of the portrait generation task, we want to retain the overall composition, pose, and appearance from the reference style image, excluding the facial identity. To achieve this, we employ the masked reconstruction objective given by:
\begin{equation}
\mathcal{L}^c_{mask} = \mathbb{E}_{z_r, c_c, \epsilon, t}\Bigr[||\epsilon \odot M_r - \epsilon_{\theta}(z_r^{(t)},t,c_c)\odot M_r ||^2_2\Bigr].
\label{eq:masked_composed}
\end{equation}
For the facial regions, we use an auxiliary identity loss that utilizes a pre-trained face recognition model~\cite{deng2019arcface} $\mathcal{R}$ and cropping function $\mathcal{B}$ conditioned by the face detection model~\cite{deng2020retinaface}:
\begin{equation}
\mathcal{L}_{id} = \mathbb{E}_{\hat{x}^{(0)}, I_s}\Bigr[1 - \text{cos}(\mathcal{R}(\mathcal{B}(\hat{x}^{(0)})), \mathcal{R}(\mathcal{B}((I_s)))\Bigr],
\label{eq:id_loss}
\end{equation}
where cos denotes the cosine similarity and $\hat{x}^{(0)}=\mathcal{D}(\hat{z}^{(0)})$ refers to the estimated clean image from $z^{(t_{id})}_r$ using Tweedie's formula~\cite{kim2021noise2score}. Timestep $t_{id}$ is sampled as $t_{id} \sim$ Uniform(1, $T^{'}$), where $T^{'} < T$, to avoid blurry and inaccurate $\hat{x}^{(0)}$ estimated from noisy latent with large timesteps, which can impair cropping or yield odd facial embeddings.

We augment the composed prompt $c_c$ by randomly selecting from predefined prompt templates to boost editing stability and generalization.

\begin{table}
\centering
{\begin{tabular}{@{}ccccc@{}}
    \toprule
     Method            &  CSIM $\uparrow$ & Style $\uparrow$     & Aesthetic $\uparrow$ \\
    \midrule
    DreamBooth         & 0.102          & 0.720                & 5.770 \\
    Textual Inversion  & 0.224          & 0.623                & 5.670 \\
    Custom Diffusion   & 0.436          & 0.606                & 5.263 \\
    \midrule
    %\hdashline

    \textbf{Ours w/o AR \& CP}     & 0.429          & 0.726                & 6.178 \\
    \textbf{Ours}      & \textbf{0.566} & \textbf{0.730}                & \textbf{6.218} \\
    \bottomrule
  \end{tabular}}
  \caption{Quantitative comparison of our method against DreamBooth~\cite{ruiz2023dreambooth}, Textual Inversion~\cite{gal2022image}, and Custom Diffusion~\cite{kumari2023multi}. Our method outperforms other baselines in terms of identity similarity measured between the source images (\textbf{CSIM}), masked CLIP similarity measure (\textbf{Style}), and \textbf{Aesthetic score}~\cite{laionaesthetic}.}
  \vspace{-0.5cm}
  \label{tab:comparison}

\end{table}

%Receptive field?? 
\noindent\paragraph{Attention Refocusing.} 
When optimizing with training images, it is vital to achieve \textbf{\textit{information disentanglement}}, ensuring that special tokens exclusively embed the information of the region of interest, denoted as $M_{v}$ for $v \in \{s,r\}$.
However, the masked reconstruction objective falls short of this goal because the presence of transformer layers in the UNet backbone gives the model a global receptive field.
%because the diffusion model's UNet backbone contains transformer layers which makes the model's receptive field global.
%In other words, masked reconstruction objective cannot penalize the special tokens from attending to the unwanted region ($\neg M_{v}$) of the given images in the optimization steps.
%This means that the masked reconstruction objective fails to penalize special tokens for attending to unwanted regions ($\neg M_{v}$) during optimization.
%Same holds for denoising steps in inference stage, where we want the attention maps of the special tokens to attend only to the intended areas. 
%For example, special token [$V1$] in the portrait generation task should only attend to the facial regions when generating images, otherwise there will be a \textbf{\textit{information spill}}.
The same limitation applies to denoising steps in the inference stage, where we desire attention maps of special tokens to focus only on the intended areas. For instance, in the portrait generation task, the special token [$V1$] should only attend to facial regions when generating images to avoid \textbf{\textit{information spill}}. 
We observe that information spill is more prevalent when the model encounters an unseen prompt during inference. 
Fig.~\ref{fig:ar} demonstrates that special tokens do indeed attend to unwanted regions.
%We find that information spill occurs more when the model is conditioned with unseen prompt in the inference stage.
%The visualization of the cross attention maps in Fig.~\ref{} shows that special tokens indeed attend to the unwanted regions.

To solve this issue, we propose a novel Attention Refocusing (AR) loss, which steers the cross attention maps $A_k$ of the special token [$V^*$] (where $k = \text{index}([\text{$V^*$}])$) using a binary target mask. 
%Our Attention Refocusing loss has two important details, the first is that the loss is applied only for $\neg M_{v}$ regions, where the mask value is 0.
Our AR loss incorporates two crucial details: First, it is applied only to regions where $\neg M_{v}$, where the mask value is zero. For the attention map values $A_k[i,j]$ where $(i,j) \in \{(i,j) | M_{v}[i,j] = 1\}$, the optimal values can vary across different UNet layers and denoising time steps, so they do not necessarily have to be close to 1. 
Conversely, for $A_k[i,j]$ where $(i,j) \in \{(i,j) | M_{v}[i,j] = 0\}$, the values should be forced to 0 to achieve information disentanglement during training and minimize information spill in the inference stage.
Second, it is essential to scale the attention maps to the [0,1] range. 
Both of these techniques are required to avoid disrupting the pre-trained transformer layers' internal operations, which would lead to corrupted outputs. The Attention Refocusing loss can be formulated as follows:
%Attention map values $A_k[i,j]$ of the $(i,j) \in \{(i,j) | M_{v}[i,j] = 1\}$ does not always have to be near 1, optimal values can vary for different UNet layers and denoising time steps.
%On the other hand, $A_k[i,j]$ of the $(i,j) \in \{(i,j) | M_{v}[i,j] = 0\}$ should be forced to 0 in order to achieve information disentanglement in training phase and reduce information spill in the inference stage.
%Second, scaling attention maps to [0,1] range is necessary. 
%Both of these techniques are required, otherwise, the internal operating mechanism of the pretrained transformer layers will break, leading to a corrupted output.
%The Attention Refocusing loss can be formulated as:
\begin{equation}
\mathcal{L}_{attn} = \mathbb{E}_{k, v \in \{s,r\}}\Bigr[||(\mathcal{S}(A_k) - M_v) \odot \neg M_v ||^2_2 \Bigr],
\label{eq:attention_refocusing}
\end{equation}
where $\mathcal{S}(\cdot)$ refers to a scaling function.

\noindent\paragraph{Postprocessing.}
The quality of images generated in a few-shot customization task is typically constrained by the capabilities of the pretrained text-to-image model used. Moreover, when provided with low-resolution source and target images, the fine-tuned model tends to produce lower-quality images. To overcome these limitations and further enhance the fidelity of the generated images, our pipeline includes optional postprocessing steps.
Specifically, we employ a pre-trained super-resolution model~\cite{wang2021realesrgan} and a face restoration model~\cite{zhou2022codeformer} to further improve the quality of the generated samples.
%The quality of the generated images generally cannot exceed the upper bound of the pretrained text-to-image model in such few-shot customization task. Also, given low-resolution source and target images, the fine-tuned model also generates low quality images. In order to get rid of these dependencies and to further boost the output fidelity, our pipeline incorporates an optional postprocessing steps.
%We use a pre-trained superresolution model~\cite{} and a face restoration model~\cite{} to further enhance generated samples.

\section{Experiments}

\begin{figure*}[htbp]
  \includegraphics[width=\linewidth]{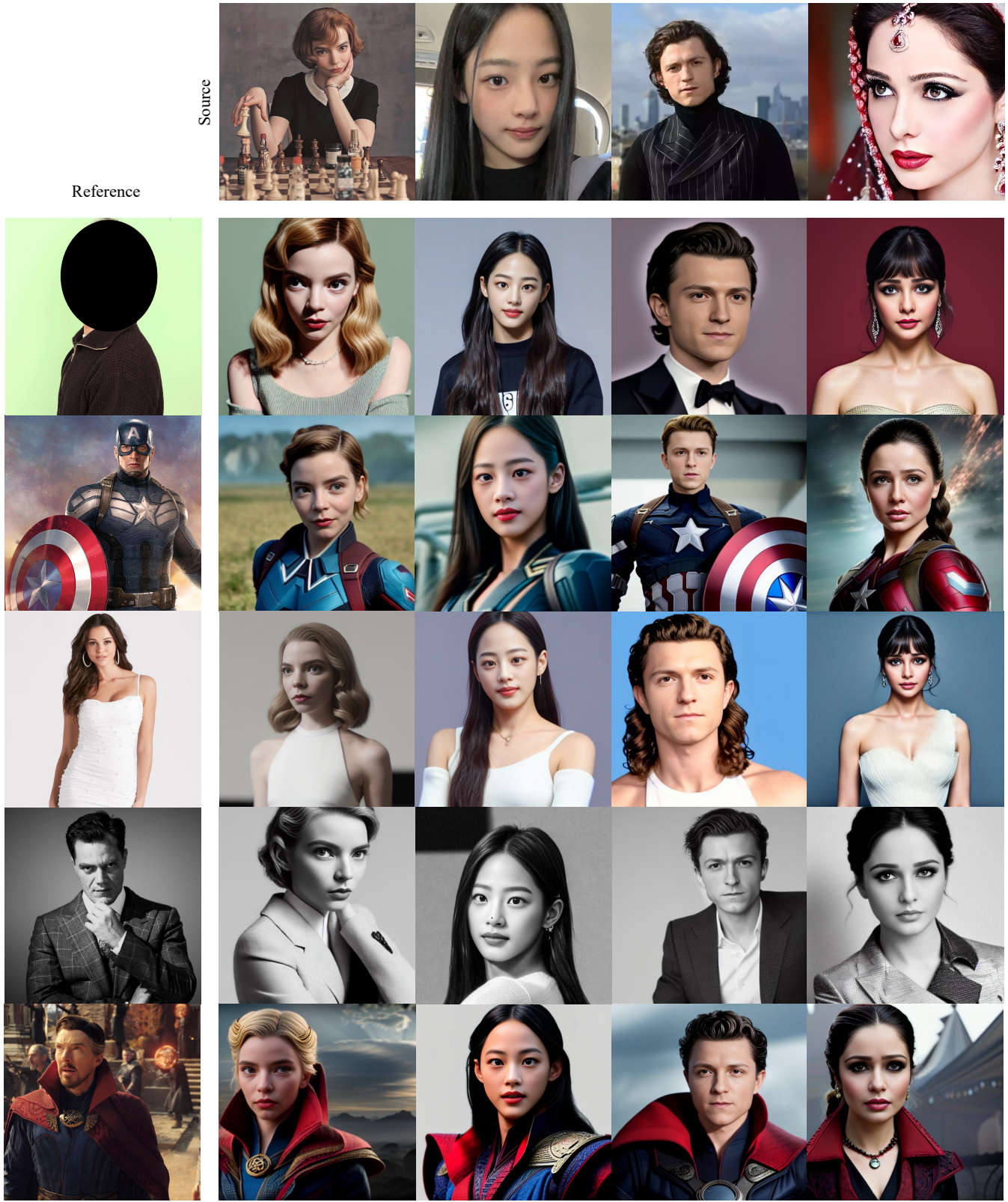}
  \caption{Curated results of MagiCapture.}
  \label{fig:curated}
  \vspace{-0.3cm}   
\end{figure*} 

\noindent\paragraph{Training Details.} Our method utilizes pre-trained Stable Diffusion V1.5~\cite{rombach2022high}. 
The first training phase consists of a total of 1200 steps, with a learning rate 5e-4 for updating the text embeddings. 
In the second LoRA phase, the learning rate is 1e-4 for the projection layers and 1e-5 for the text embeddings, with a total of 1500 training steps. 
The model is trained on a single GeForce RTX 3090 GPU, using a batch size of 1 and gradient accumulation over 4 steps. 
For all experiments, we employ 4 to 6 images for both the source and reference images. Please refer to the supplement for more details.

\begin{figure}[t!]
\centering
\includegraphics[width=1.05\linewidth]{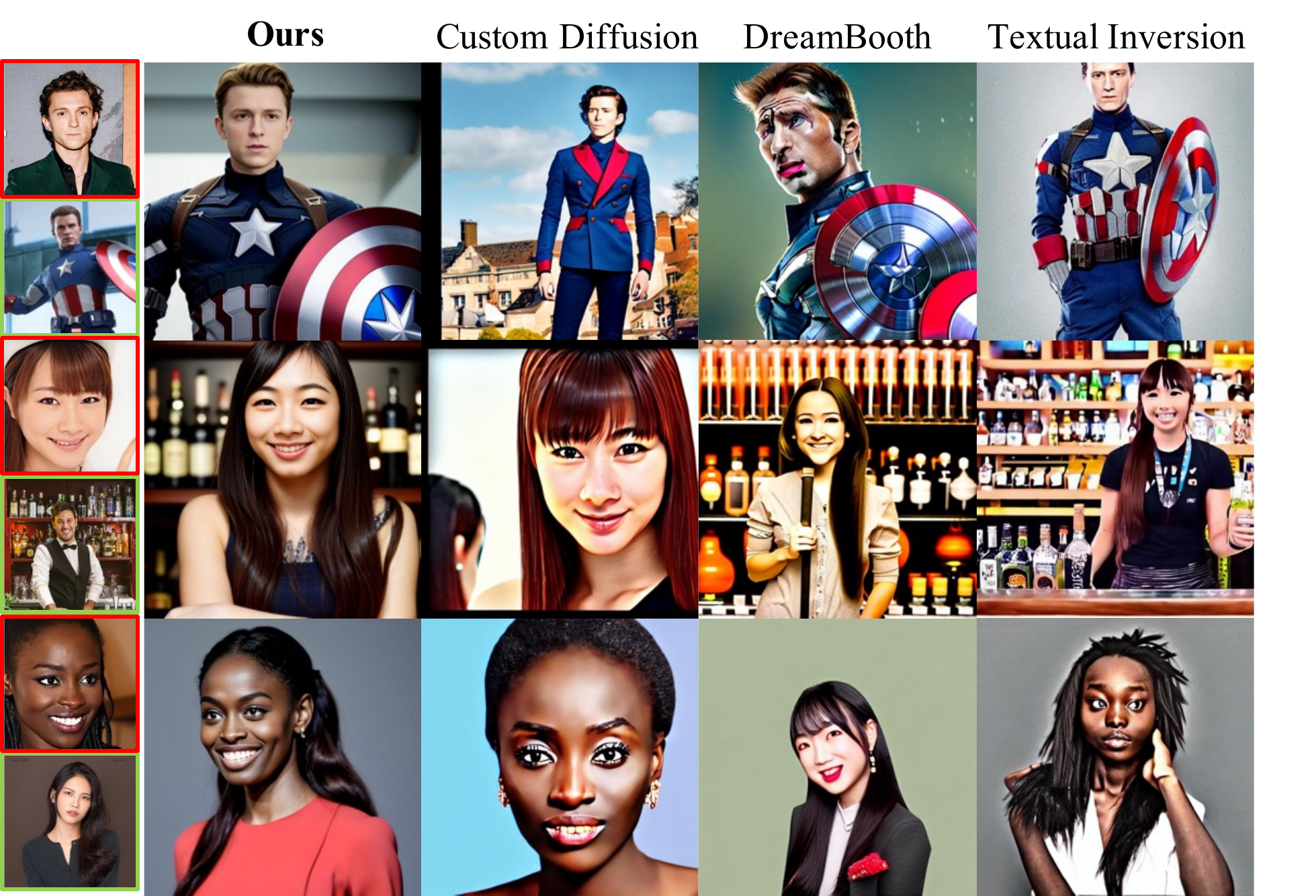}
%\vspace{-1.5em}
\caption{Qualitative comparisons of MagiCapture with other baseline methods.}
\label{fig:qual}
%\vspace{-1.0em}
\end{figure}

\noindent\paragraph{Comparisons.}
The results of our method are demonstrated in Fig.~\ref{fig:curated}.
We compare our method with other personalization methods including DreamBooth~\cite{ruiz2023dreambooth}, Textual Inversion~\cite{gal2022image}, and Custom Diffusion~\cite{kumari2023multi} using the same source and reference images.
We choose 10 identities, 7 from VGGFace~\cite{cao2018vggface2} and 3 in-the-wild identities gathered from the internet.
We also manually select 10 style concepts, leading to 100 id-style pairs.
For each pair, we train each baseline and our model, then generate 100 images with the \textit{composed prompt} for each of the trained model, resulting in 10,000 samples per baseline.
Qualitative comparisons are shown in Fig.~\ref{fig:qual}, where our method outperforms other baselines in image fidelity and source-reference image reflection.

We assess the facial appearance similarity between the source and generated portrait images by measuring the cosine similarity between their facial embeddings, using a pre-trained recognition network (CSIM)~\cite{zakharov2019few}.

Another important aspect of evaluation is style preservation, where we measure how well the results replicate the style of the reference images.
%Another important aspect to evaluate is style preservation, where the goal is to measure how well the results capture the style of the reference images.
We compute the cosine similarity between the \textit{masked} CLIP~\cite{radford2021learning} image embeddings of the reference and generated images, where facial regions are masked to exclude facial appearance from the assessment.
%We measure cosine similarity between \textit{masked} CLIP~\cite{radford2021learning} image embeddings of the reference and generated images, where the facial regions are masked out so that the facial appearance is ignored in the evaluation.
We use CLIP similarity instead of texture similarity~\cite{gatys2016image} since the term \textit{style} in our paper encompasses broader concepts such as image geometry and composition, in addition to texture and appearance of non-facial regions.
%in our context incorporates wider meaning including geometry and composition of the image as well as texture and appearance of non-facial regions.
Finally, we evaluate the overall image fidelity with the LAION aesthetic predictor~\cite{laionaesthetic}.
Table~\ref{tab:comparison} shows that our method outperforms other baselines in all three metrics.
%We also conduct a user study to complement the above metrics. 
%User study is conducted with 30 people and asked to rate the score in terms of ID preservation, style preservation, and image fidelity, in the range of 1-5. 
Additionally, we conduct a user study involving 30 participants who were asked to rate images for ID preservation, style preservation, and image fidelity on a 1-5 scale.
Table~\ref{tab:Userstudy} summarizes the results, with our method consistently scoring higher than other baselines.
%Summary of the study is introduced in Table~\ref{tab:Userstudy}, where our method shows better scores compared to other baselines by a large margin.

%We observe that DreamBooth frequently overfits the reference style images, which explains the reason for the high style score but low CSIM.
%On the other hand, textual inversion generally underfits to both of the source and the reference images, resulting in low-fidelity images that do not preserve the appearance details of the source and the reference.
%Custom Diffusion better preserves the source identity compared to the two, but the method still cannot guarantee its performance for the composed prompt, which leads to identity shift and unnatural images.
We observed that DreamBooth often overfits to the reference style images, leading to high style scores but low CSIM scores.
Conversely, Textual Inversion tends to underfit both the source and reference images, resulting in low-fidelity images that fail to preserve appearance details.
Custom Diffusion better preserves source identity compared to the others, but still cannot consistently perform well for the composed prompt, leading to identity shifts and unnatural images.

\begin{table}
\centering
  {\begin{tabular}{@{}ccccc@{}}
    \toprule
     Method            &  ID $\uparrow$ & Style $\uparrow$     & Fidelity $\uparrow$ \\
    \midrule
    DreamBooth         &    2.025       &    3.648             & 2.683 \\
    Textual Inversion  &     2.907  &       3.038          &  2.965\\
    Custom Diffusion   &  3.223         &   2.260              &  2.980 \\
    \textbf{Ours}      & \textbf{4.055} &    \textbf{4.165}             & \textbf{4.293} \\
    \bottomrule
  \end{tabular}}
  \caption{User study of our method against DreamBooth~\cite{ruiz2023dreambooth}, Textual Inversion~\cite{gal2022image}, and Custom Diffusion~\cite{kumari2023multi}. Our method outperforms other baselines in terms of identity similarity score (\textbf{ID}), style similarity measure (\textbf{Style}), and image fidelity score (\textbf{Fidelity}).}
  \vspace{-0.3cm}
  \label{tab:Userstudy}
\end{table}

\noindent\paragraph{Ablation Study.} As shown in Fig.~\ref{fig:ar}, we find that Attention Refocusing loss effectively prevents attention maps from attending to unwanted regions, mitigating information spill and promoting information disentanglement.
Empirically, we observe that the Attention Refocusing loss should only be applied during the second phase of training (LoRA training).
We infer that text embeddings are not well-suited for learning geometric information related to attention maps.
Moreover, without composed prompt learning, the generated images often exhibit undefined behaviors where only one of the source or reference sets is evident in the image, without blending.
%We empirically find that Attention Refocusing loss should only be applied in the second phase of the training (LoRA training).
%We hypothesize that text embeddings are not adequate to learn geometric information that is related to attention maps.
%Also, without composed prompt learning, the results frequently display undefined behaviors where only one of the source or the reference sets are reflected to the image, unmixed.
We present the evaluation metrics for both the presence and absence of composed prompt learning (CP) and Attention Refocusing (AR) in Table~\ref{tab:comparison}.
%We report the assessment measures with and without composed learning (CP) and Attention Refocusing (AR) in Table~\ref{tab:comparison}. 
For more results and detailed analysis, please refer to the supplement.

\begin{figure}[t!]
\centering
\includegraphics[width=1.2\linewidth]{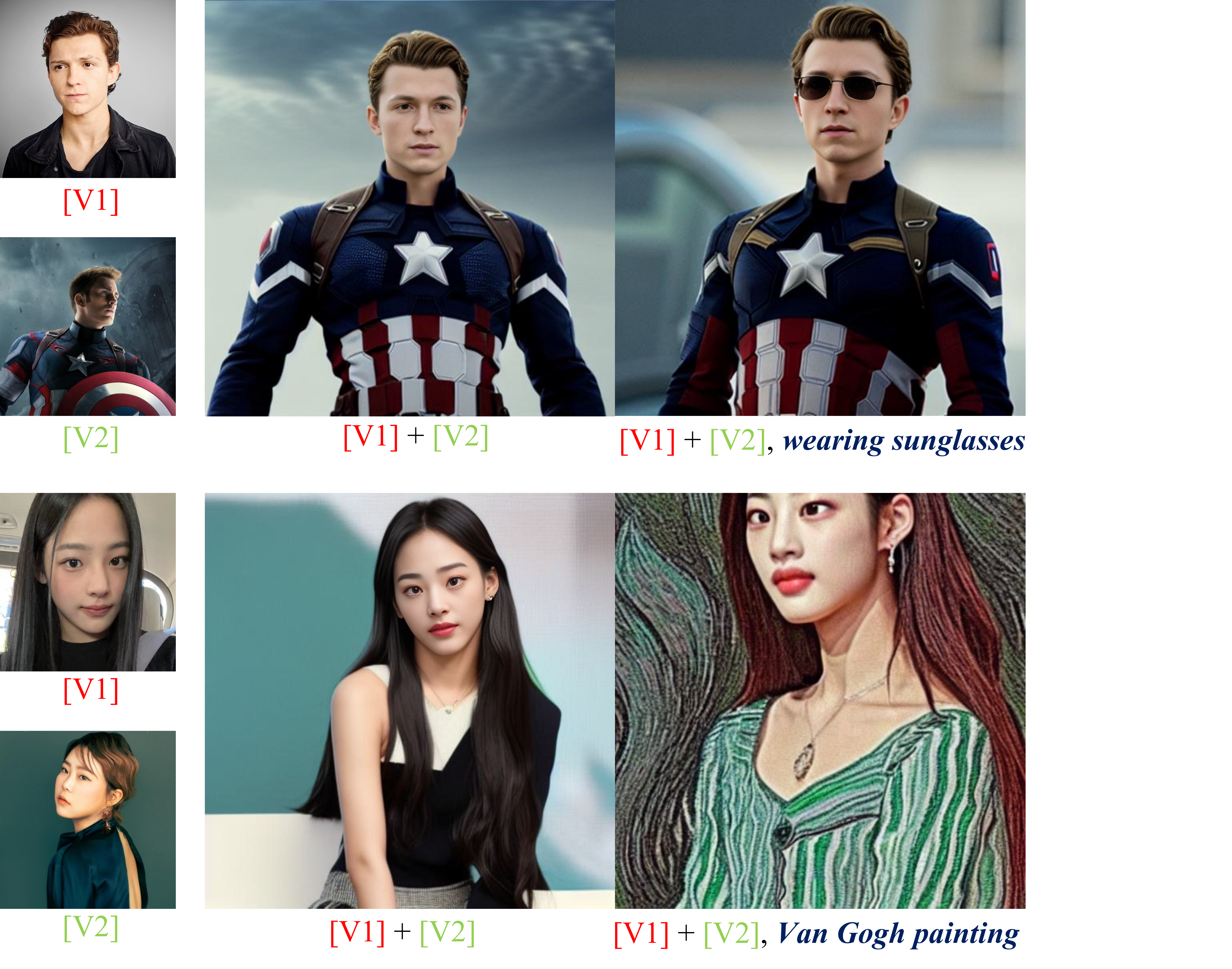}
\vspace{-0.5cm}
\caption{Users can further manipulate the composed results using prompts with additional description.}
\label{fig:editing}
\vspace{-0.4cm}
\end{figure}

\noindent\paragraph{Applications.} Since our method is robust to generalizations, users can further manipulate the composed results using prompts with more descriptions (e.g., $c_c^{'} =$ ``A photo of [$V1$] person in the [$V2$] style, wearing sunglasses.").
We demonstrate such results in Fig.~\ref{fig:editing} and in the supplement.
Furthermore, our method is adaptable for handling different types of content, including non-human images. For methodologies and results related to non-human content, please refer to the supplementary material.

\begin{figure}[htbp]
  \includegraphics[width=\linewidth]{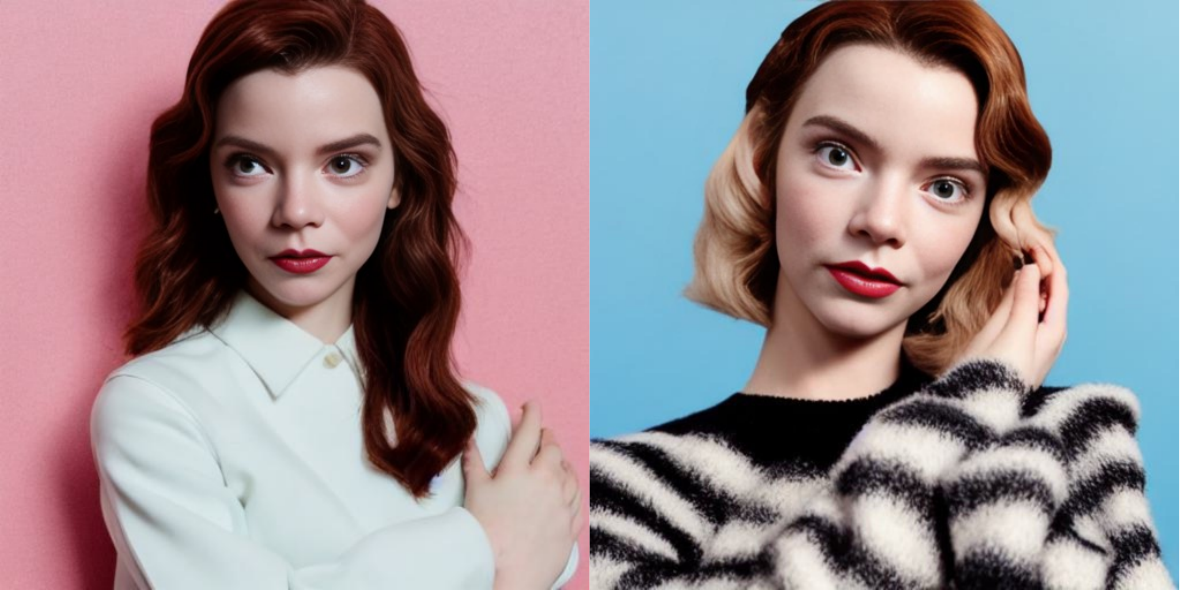}
  \caption{Failure cases: Proposed method occasionally produces abnormal body parts such as limbs, fingers}
  \label{fig:fail}
  \vspace{-0.3cm}
\end{figure} 

\section{Limitations and Conclusions}
Our method occasionally produces abnormal body parts such as limbs, fingers, as shown in Fig.~\ref{fig:fail}.
Furthermore, the model tends to exhibit lower fidelity for non-white subjects and demonstrates a noticeable gender bias—for instance, it struggles to accurately generate images of men wearing wedding dresses.
These issues are largely related to the inherent biases of the pre-trained text-to-image models, and addressing these problems within a few-shot setting represents a significant avenue for future research. 
We acknowledge the ethical implications of our work and are committed to taking them seriously. We are also proactive in leading and supporting efforts to prevent potential misuse of our contributions.

\section{Acknowledgements}
This work was supported by the National Research Foundation of Korea (NRF) grant funded by the Korea government (MSIT) (No. NRF-2022R1A2B5B02001913), and Institute of Information \& communications Technology Planning \& Evaluation (IITP) grant funded by the Korea government (MSIT) (No.2019-0-00075, Artificial Intelligence Graduate School Program (KAIST)).

\bibliography{MagiCapture}

\clearpage

\section{Supplementry Materials}
\noindent 
\noindent

\subsection{Training Details}
\noindent \paragraph{MagiCapture}
The loss function for the first phase training is given as:
\begin{equation}
\mathcal{L}^s_{mask} + \mathcal{L}^r_{mask}.
\end{equation}
For the second phase LoRA training, composed prompt learning and Attention Refocusing loss is added:
\begin{equation}
\mathcal{L}^s_{mask} + \mathcal{L}^r_{mask} + \mathcal{L}^c_{mask} + \lambda_{id}\mathcal{L}_{id} + \lambda_{attn}\mathcal{L}_{attn},
\end{equation}
where $\lambda_{id} = 1$ and $\lambda_{attn} = 2.5$ is used for all experiments. For $\lambda_{id}$, 0.25 or 0.5 are also fine.

\noindent \paragraph{DreamBooth} 
%For DreamBooth~\cite{ruiz2023dreambooth} training, we utilize the best setting with prior preservation, using lambda value of 1.0 and 200 imgs.
%Batch size is 2, including 1 source and 1 style image for each batch.
%The learning rate is 1e-6 and is trained for 2000 steps, and CLIP text encoder is trained along with the UNet parameters.
%For inference, denoising step of 50 with guidance scale 7.5 is used.
We employ the optimal settings for DreamBooth~\cite{ruiz2023dreambooth} training, which include prior preservation with a lambda value of 1.0 and a dataset of 200 images. Each batch comprises two images, consisting of one source and one style image. We set the learning rate to 1e-6 and train the model for 2000 steps. During this training, the CLIP text encoder and the UNet parameters are concurrently optimized. For inference, we use a denoising step of 50 with a guidance scale of 7.5.

\noindent \paragraph{Custom Diffusion}
%For Custom Diffusion~\cite{kumari2023multi} training, we utilize the best setting with prior preservation, using lambda value of 1.0 and 200 imgs.
%Batch size is 2, and the learning rate is 5e-6 and is trained for 750 steps, optimizing cross attention layers of the Stable Diffusion~\cite{rombach2022high} as introduced in the paper.
%For inference, denoising step of 50 with guidance scale 7.5 is used.
For training the Custom Diffusion model~\cite{kumari2023multi}, we use the best settings with prior preservation, a lambda value of 1.0, and a dataset of 200 images. The batch size is set to 2. With a learning rate of 5e-6, we train the model for 750 steps, optimizing the cross-attention layers of the Stable Diffusion model~\cite{rombach2022high}, as detailed in the original paper. The inference phase employs a denoising step of 50 and a guidance scale of 7.5.

\noindent \paragraph{Textual Inversion}
%For Textual Inversion~\cite{gal2022image} training, we utilize the best setting with batch size 2, the learning rate of 5e-3 and total of 5000 training steps.
%For inference, denoising step of 50 with guidance scale 7.5 is used.
For the training of Textual Inversion~\cite{gal2022image}, we adopt the optimal settings, including a batch size of 2, a learning rate of 5e-3, and a total of 5000 training steps. The inference process involves a denoising step of 50 with a guidance scale of 7.5.

\begin{figure}[h]
  \includegraphics[width=\linewidth]{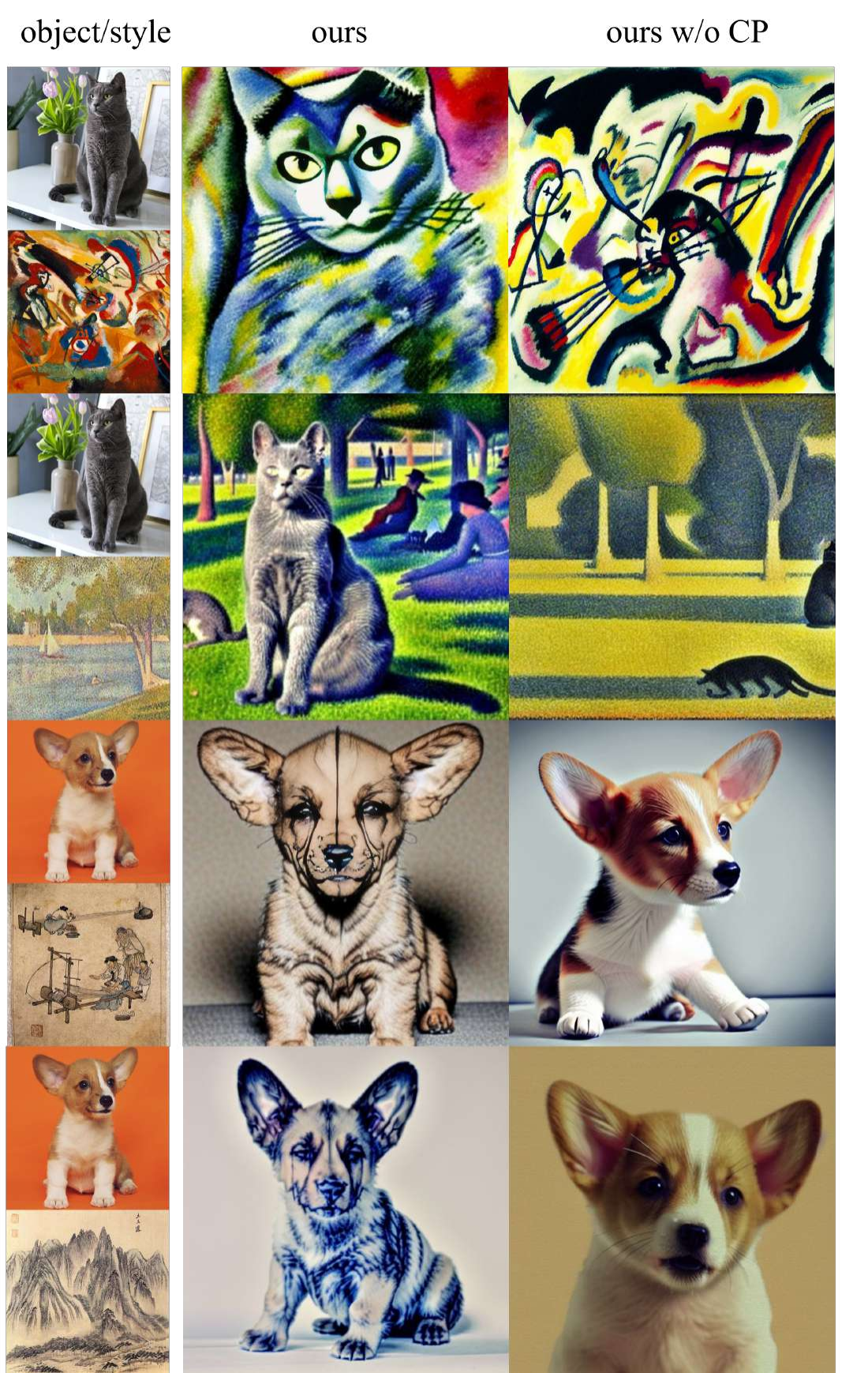}
  \caption{A comparison with results produced without the use of composed prompt learning for non-human images.}
  \label{fig:sup_general_ab}
\end{figure} 

\subsection{General Object}
Our method can also be applied to other general objects, where our composed prompt learning can be applied for robust multi-concept composition.
%We demonstrate a case where we want to retain the structure of the source object and apply texture of the reference image.
We illustrate this with an example where the goal is to maintain the structure of the source object while adopting the texture from the reference image. 
We employ the same masked reconstruction objective $\mathcal{L}^s_{mask}$ for the source, and naive reconstruction objective without masking $\mathcal{L}^r$ for the reference.

For composed prompt learning, we employ structure loss~\cite{kwon2022diffusion} that maximizes structural similarity between the estimated image $\hat{x}^{(0)}$ and the source images using a pre-trained DINO ViT~\cite{caron2021emerging}.
Specifically, the structure loss comprises two components: the self-similarity loss $\mathcal{L}_{ssim}$~\cite{tumanyan2022splicing} and the patch contrastive loss $\mathcal{L}_{contra}$~\cite{park2020contrastive}.
$\mathcal{L}_{ssim}$ utilizes a self similarity matrix derived from the multi-head self attention (MSA) layer of the pre-trained DINO.
$\mathcal{L}_{contra}$ maximizes the patch-wise similarity between the keys of the source and the estimated image $\hat{x}^{(0)}$, with the keys extracted from the MSA layer of DINO.
For the style similarity loss $\mathcal{L}_{style}$, we minimize the distance between DINO ViT [CLS] token embeddings of the reference and the estimated image $\hat{x}^{(0)}$.
To sum up, our loss function for composed prompt learning is:
\begin{equation}
\lambda_{ssim}\mathcal{L}_{ssim} + \lambda_{contra}\mathcal{L}_{contra} + \lambda_{style}\mathcal{L}_{style},
\end{equation}
where $\lambda_{ssim}=0.1$, $\lambda_{ssim}=0.2$, and $\lambda_{ssim}=2$ is used for training.

\begin{figure*}[htbp]
  \vspace{-0.7cm}   
  \includegraphics[width=\linewidth]{
  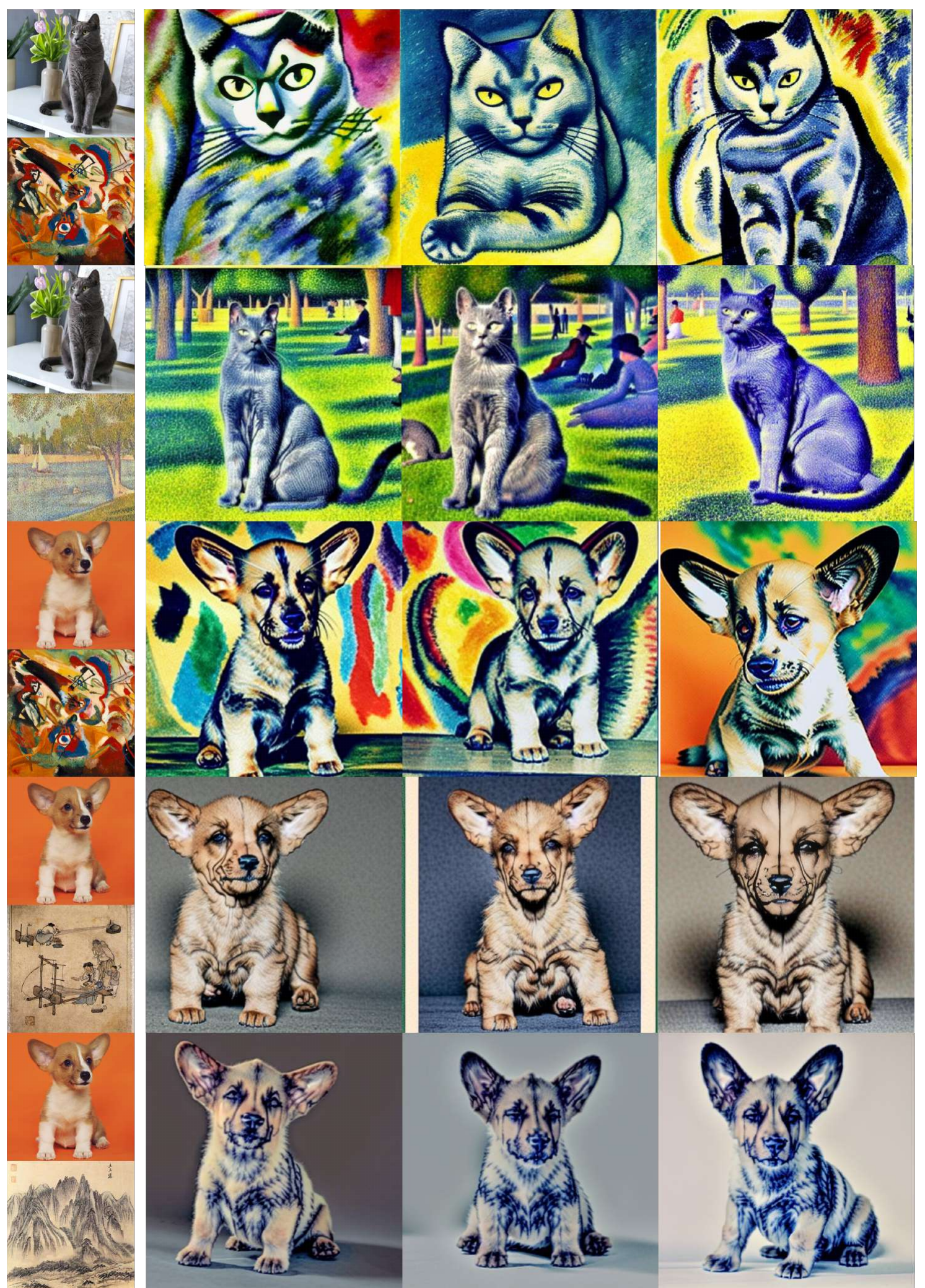}
  \caption{Results for composing the source content and the reference style in non-human images.}
  \label{fig:sup_general}
  \vspace{-0.3cm}   
\end{figure*}

We demonstrate the results for general objects in Fig.~\ref{fig:sup_general}.
Additionally, in Fig.~\ref{fig:sup_general_ab}, we provide a comparison with results produced without the use of composed prompt learning. 
These comparisons reveal that, in the absence of composed prompt learning, the outcomes tend to suffer from two main issues: either the structure of the source concept is inadequately preserved, or the style of the reference images is not effectively incorporated.

\begin{table}
\centering
{\begin{tabular}{@{}ccccc@{}}
    \toprule
     Method                     & CSIM $\uparrow$     & Style $\uparrow$     & Aesthetic $\uparrow$ \\
    \midrule
    Ours                        & \textbf{0.566}      & 0.730                & \textbf{6.218}                \\
    Ours w/ postprocessing      & 0.508               & \textbf{0.737}       & 6.184                \\
    Ours w/o CP                 & 0.429               & 0.717                & 6.159                \\
    %Ours w/o AR                 & 0.559               & 0.748                & 6.378                \\
    Ours w/o AR \& CP           & 0.429               & 0.726                & 6.178                \\
    \bottomrule
  \end{tabular}}
  \caption{The results of the ablation study clearly highlights significance of composed prompt learning (CP) in enhancing the metrics. When CP is not included, there is a noticeable decline in CSIM and style score (measured by masked CLIP similarity).}
  %\vspace{-0.5cm}
  \label{tab:sup_ablation}

\end{table}

\begin{figure}[h]
  %\vspace{-1.3cm}   
  \includegraphics[width=\linewidth]{
  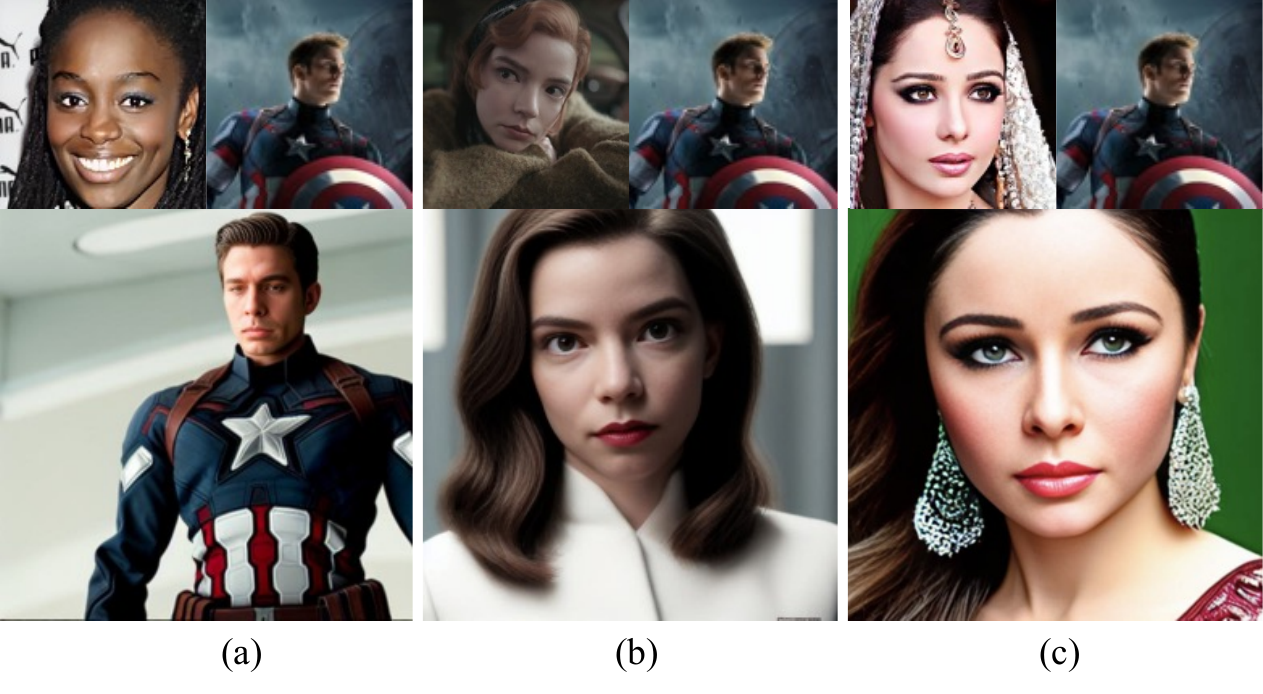}
  \caption{Results without Attention Refocusing (AR) loss. While AR loss does not appear to contribute to the metric improvement, the absence of AR often leads to \textit{collapsed} samples as seen in (a) and (b). The generated samples predominantly reflect either the source or reference images, rather than a balanced combination of both. (c) illustrates that without AR, information spill is evident in the generated earrings, indicating that the source special tokens attend to non-facial regions.}
  \label{fig:sup_ablation}
  \vspace{-0.6cm}   
\end{figure}

\subsection{Ablation Study}
\begin{figure}[H]
  \includegraphics[width=\linewidth]{
  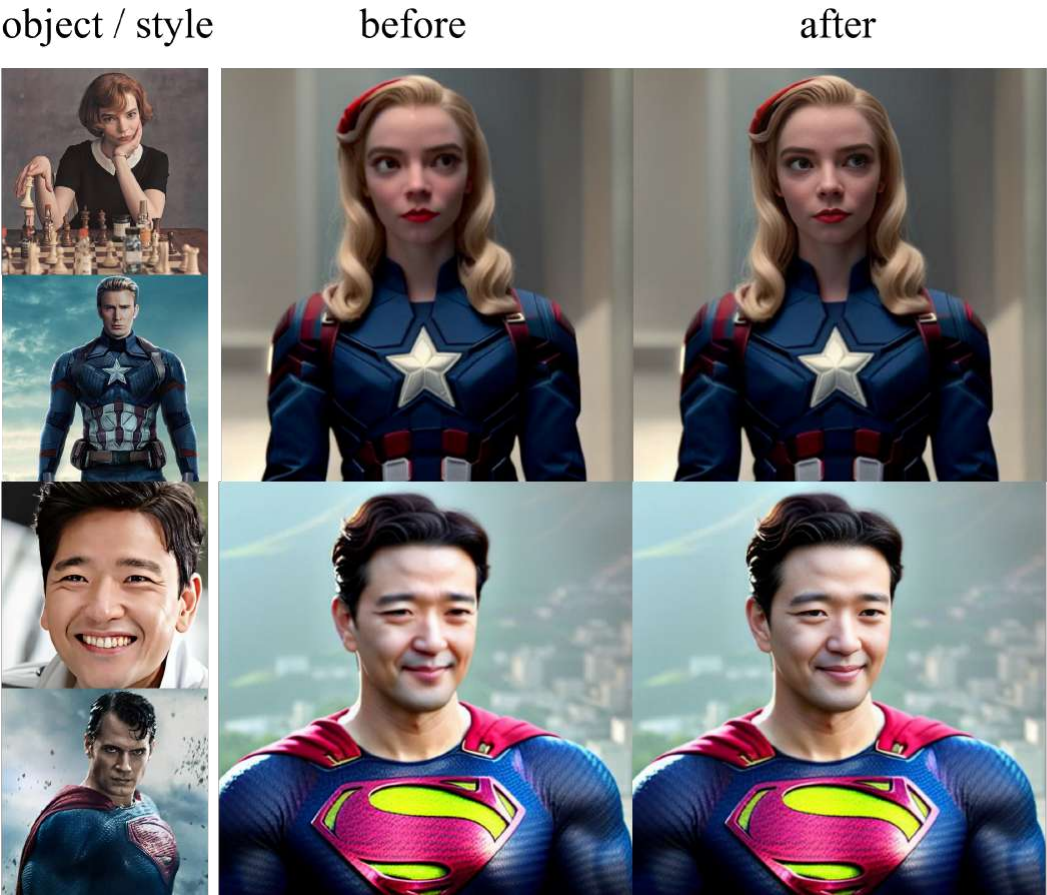}
  \caption{Generated results before and after post-processing.}
  \label{fig:sup_post}
  \vspace{-0.3cm}   
\end{figure}

We present the results of our ablation study in Table~\ref{tab:sup_ablation}, which clearly highlight the significance of composed prompt learning (CP) in enhancing the metrics. 
When CP is not included, there is a noticeable decline in CSIM and style score (measured by masked CLIP similarity). 
Conversely, while the Attention Refocusing (AR) loss does not appear to contribute to the metric improvement, it is noteworthy that the absence of AR often leads to \textit{collapsed} samples, where the generated samples predominantly reflect either the source or reference images, rather than a balanced combination of both.
Illustrative examples of this are provided in Fig.~\ref{fig:sup_ablation}, where Fig.~\ref{fig:sup_ablation} (a) showcases results that lean heavily towards the reference images, while Fig.~\ref{fig:sup_ablation} (b) exhibits only the source identity.
Additionally, we observed instances of information spill when AR loss is not applied. Fig.~\ref{fig:sup_ablation} (c) illustrates that without AR, information spill is evident in the generated earrings, indicating that the source special tokens attend to non-facial regions.
Finally, we note that the CSIM score exhibits a minor decline following post-processing. Although the post-processed results are generally visually appealing, the face restoration model possesses a level of freedom that can occasionally lead to a slight reduction in the similarity score. The results of samples before and after applying the post-processing are displayed in Fig.~\ref{fig:sup_post}.

%We demonstrate ablation results in Table~\ref{tab:sup_ablation}.
%It shows that composed prompt learning (CP) is the most important factor in improving the metrics.
%There is a large drop of CSIM and style (masked CLIP similarity) without composed prompt learning.
%On the other hand, while it seems that Attention Refocusing (AR) loss does not contribute much to the metric, we observe that generated samples often collapsed samples, where only either one of the source or reference images are reflected.
%For instance, Fig.~\ref{fig:sup_ablation} represents images generated without Attention Refocusing loss. 
%Fig.~\ref{fig:sup_ablation} (a)
%we observe generated samples with facial regions

\subsection{Curated Results}
\begin{figure*}[htbp]
  \vspace{-0.7cm}   
  \includegraphics[width=\linewidth]{
  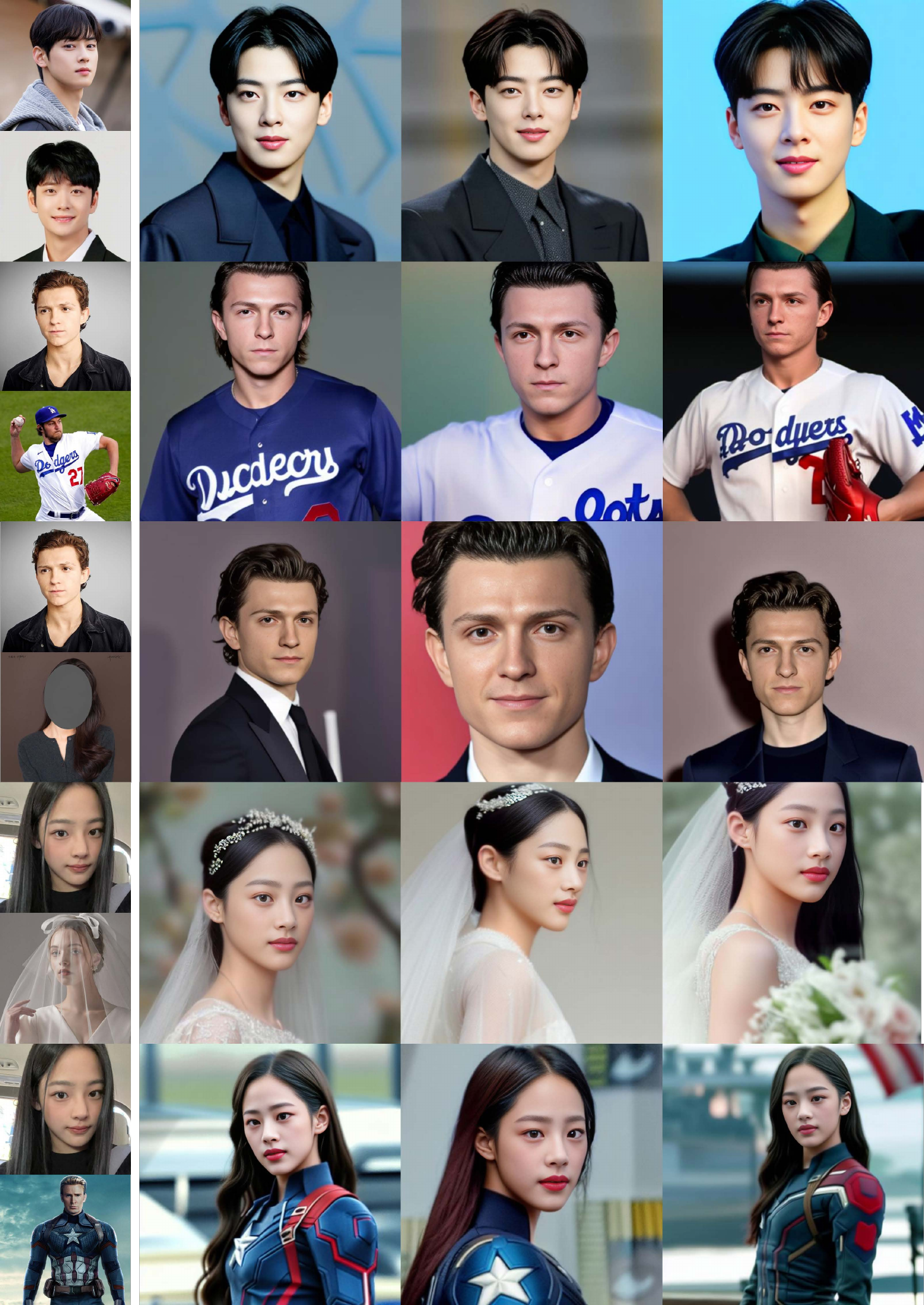}
  \caption{Curated results of MagiCapture.}
  \label{fig:sup_curated1}
  \vspace{-0.3cm}   
\end{figure*}

\begin{figure*}[htbp]
  \vspace{-0.7cm}   
  \includegraphics[width=\linewidth]{
  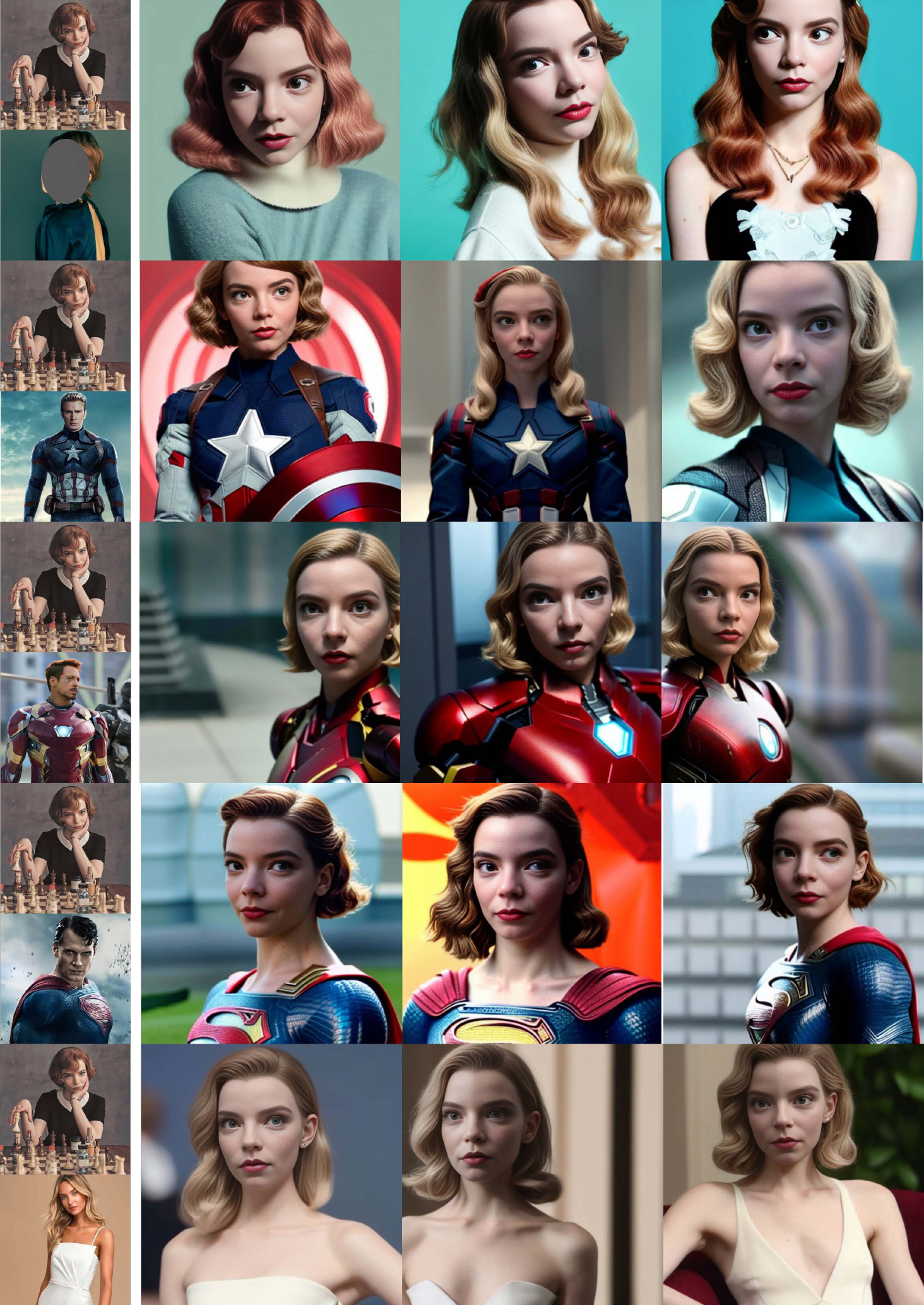}
  \caption{Curated results of MagiCapture.}
  \label{fig:sup_curated2}
  \vspace{-0.3cm}   
\end{figure*}

\begin{figure*}[htbp]
  \vspace{-0.7cm}   
  \includegraphics[width=\linewidth]{
  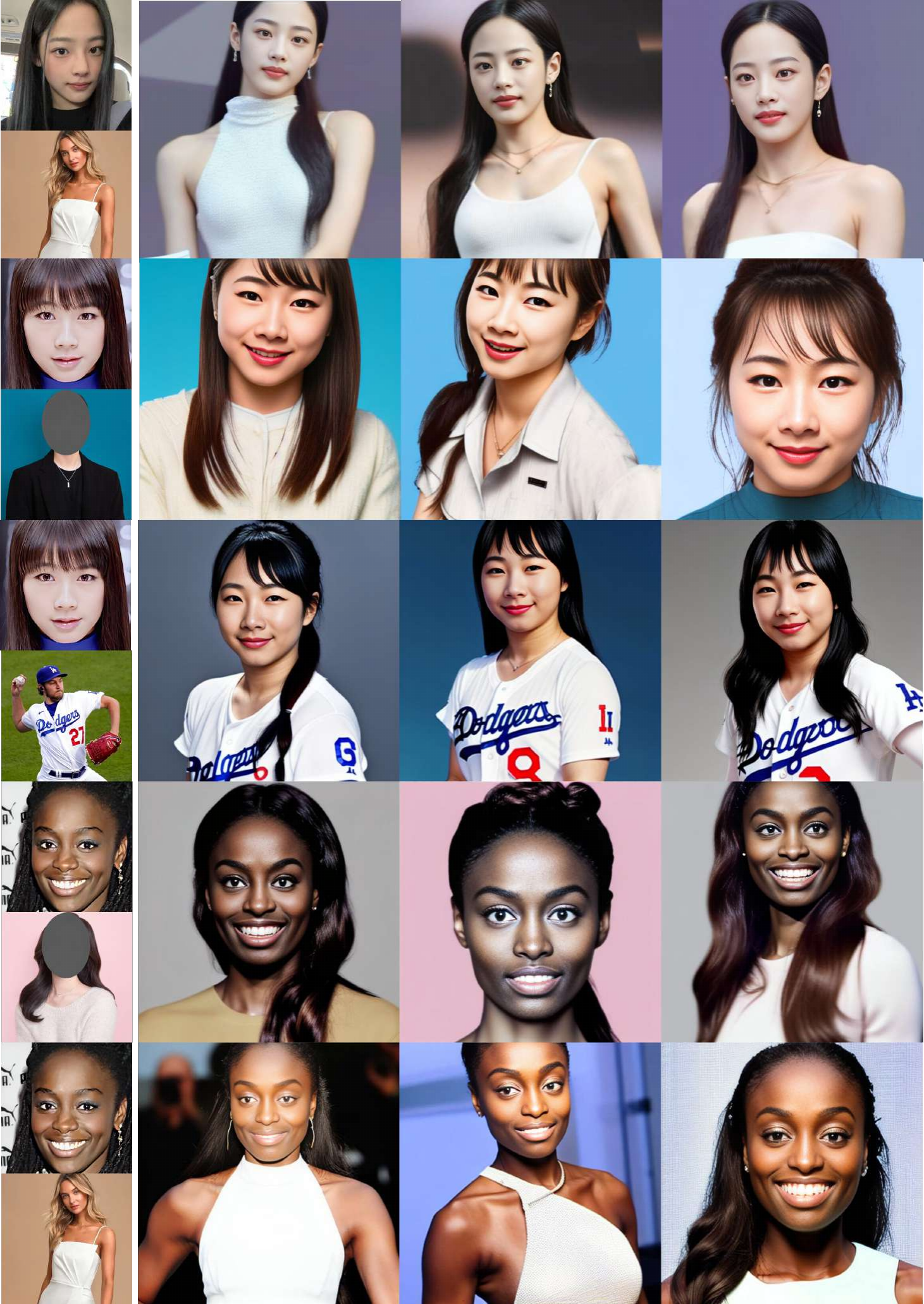}
  \caption{Curated results of MagiCapture.}
  \label{fig:sup_curated3}
  \vspace{-0.3cm}   
\end{figure*}

\begin{figure*}[htbp]
  \vspace{-0.7cm}   
  \includegraphics[width=\linewidth]{
  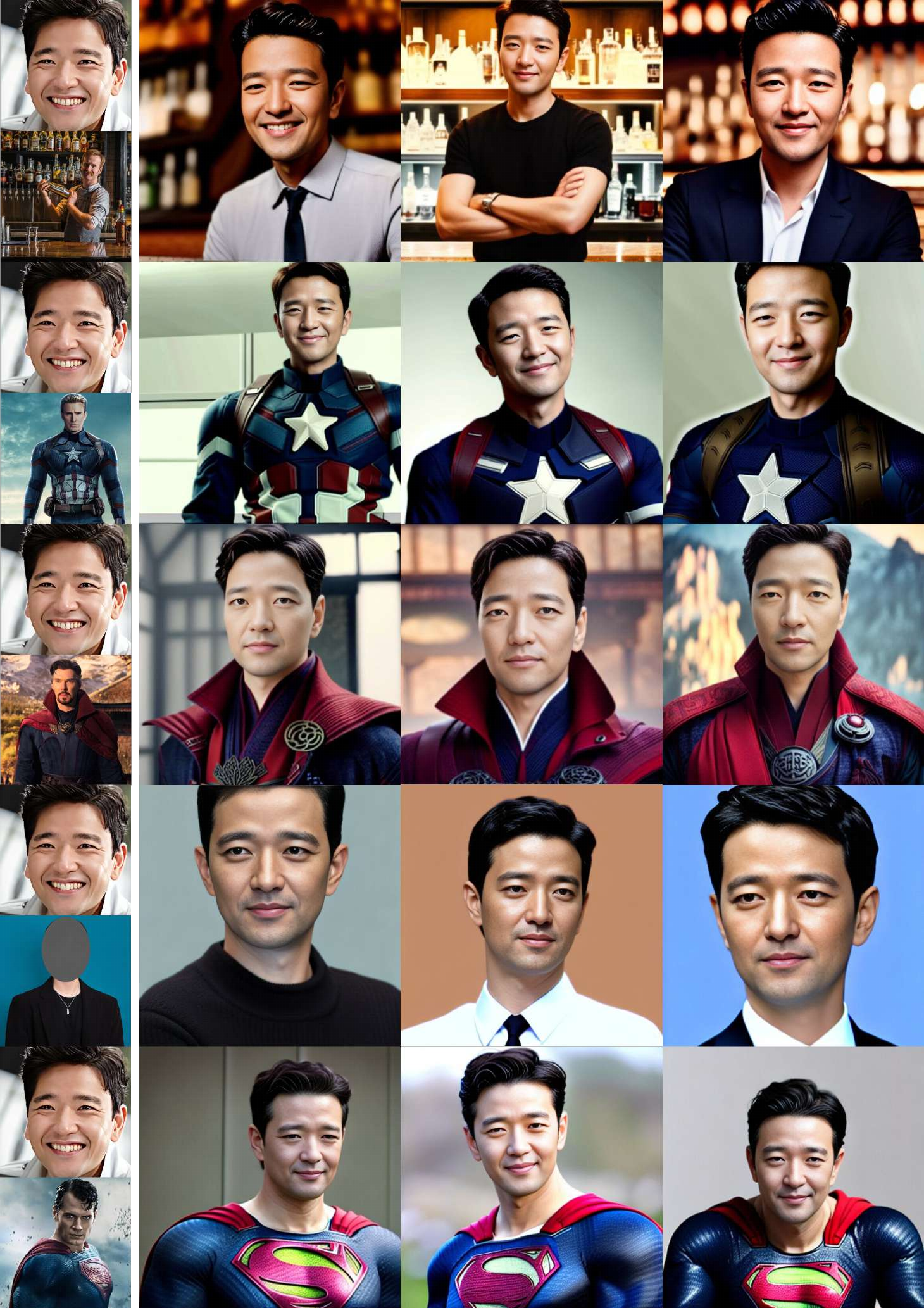}
  \caption{Curated results of MagiCapture.}
  \label{fig:sup_curated4}
  \vspace{-0.3cm}   
\end{figure*}

\begin{figure*}[htbp]
  \vspace{-0.7cm}   
  \includegraphics[width=\linewidth]{
  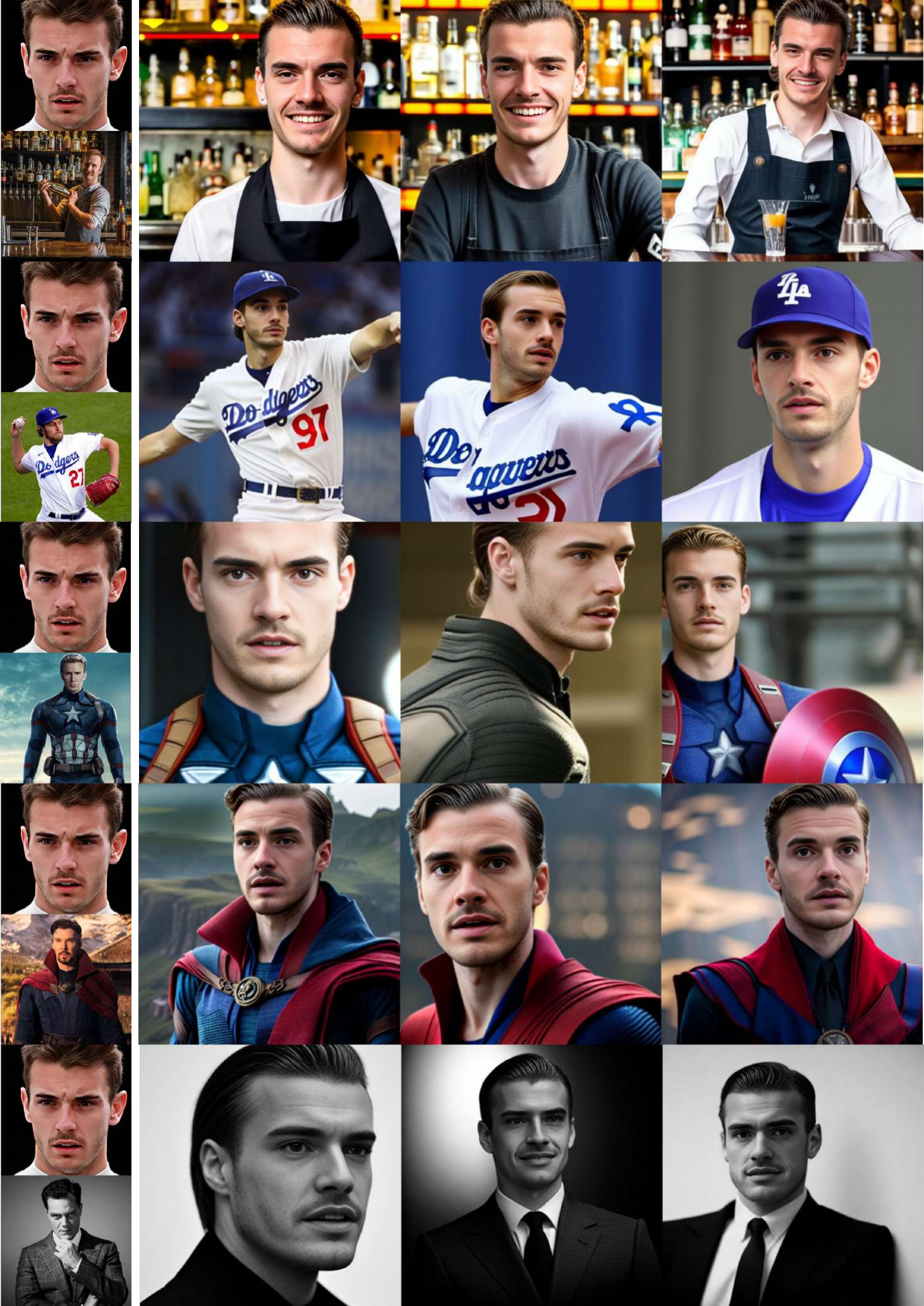}
  \caption{Curated results of MagiCapture.}
  \label{fig:sup_curated5}
  \vspace{-0.3cm}   
\end{figure*}

\begin{figure*}[htbp]
  \vspace{-0.7cm}   
  \includegraphics[width=\linewidth]{
  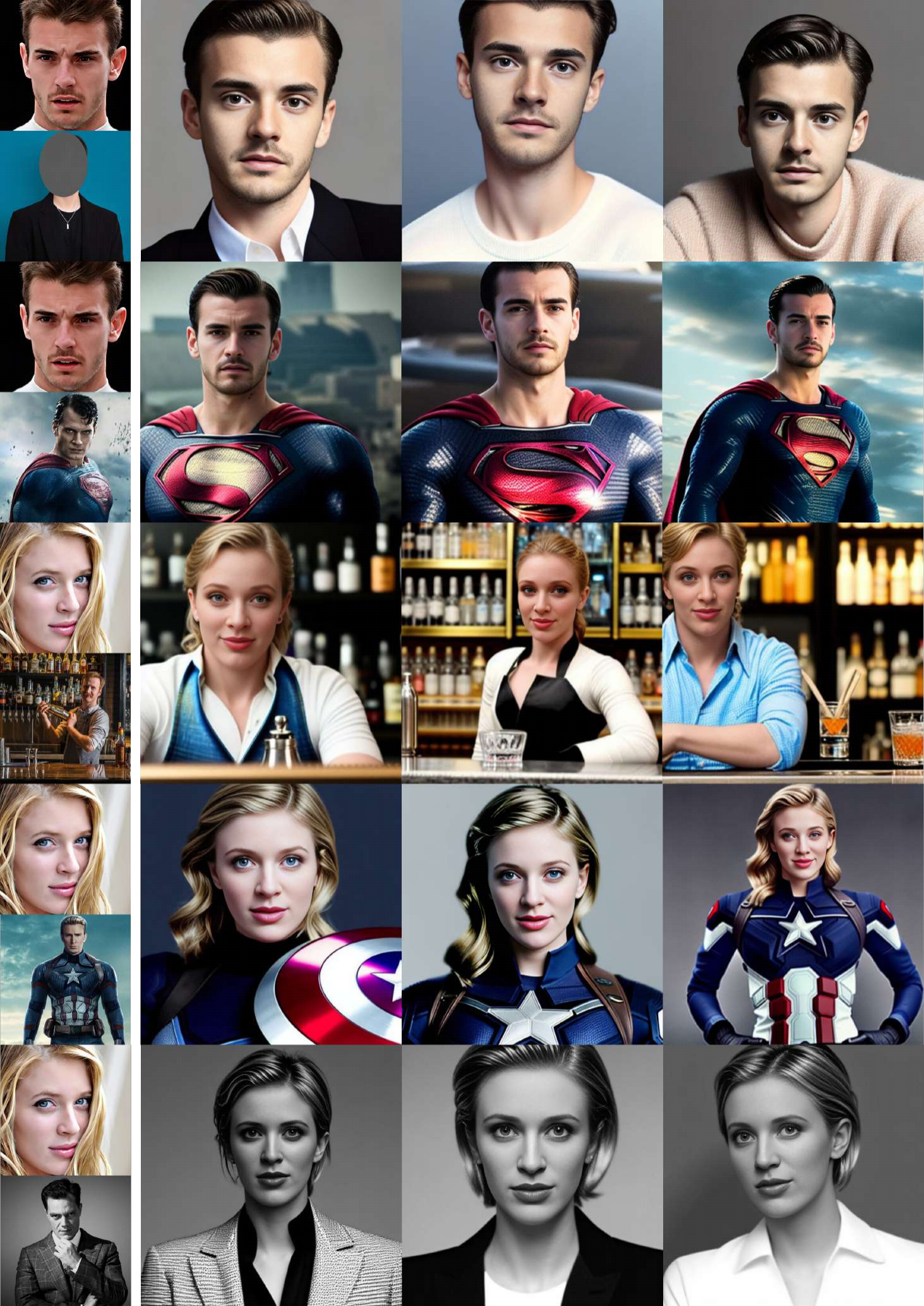}
  \caption{Curated results of MagiCapture.}
  \label{fig:sup_curated6}
  \vspace{-0.3cm}   
\end{figure*}

We demonstrate more results from Fig.~\ref{fig:sup_curated1} to Fig.~\ref{fig:sup_curated6}.

\end{document}